\documentclass[conference,10pt]{IEEEtran}
\IEEEoverridecommandlockouts
\usepackage{cite}
\usepackage{amsmath,amssymb,amsfonts}
\usepackage{mathtools}
\usepackage{algorithmic}
\usepackage{graphicx}
\usepackage{textcomp}
\usepackage{xcolor}
\usepackage{pgf, tikz, pgfplots}
\usepgfplotslibrary{groupplots}
\usetikzlibrary{shapes, arrows.meta, calc, matrix, patterns, positioning}
\usepackage{mleftright}
\usepackage{bm}
\usepackage{bbm}
\usepackage{todonotes}
\usepackage{acronym}

\usepackage[ruled,vlined, linesnumbered]{algorithm2e}
\let\oldnl\nl%
\newcommand{\nonl}{\renewcommand{\nl}{\let\nl\oldnl}}%

\SetCommentSty{mycommfont}

\newcommand{\bunderline}[1]{\mkern2mu\underline{\mkern-2mu#1\mkern-6mu}\mkern6mu }
\newcommand{\dunderline}[1]{\mkern6mu\underline{\mkern-10mu#1\mkern-9mu}\mkern6mu }
\newcommand{\xt}{u^\textrm{t}}
\newcommand{\xf}{u^\textrm{f}}
\newcommand{\WDFT}{\bm{W}}

\newcommand{\WDFTreal}{\bunderline{\bm{W}}}
\newcommand{\WIDFTreal}{\bunderline{\bm{W}}^{-1}}
\newcommand{\xtreal}{\underline{\bm{u}}^\textrm{t}}
\newcommand{\xfreal}{\underline{\bm{u}}^\textrm{f}}

\definecolor{KITgreen}{rgb}{0,.59,.51}
\definecolor{KITpalegreen}{RGB}{130,190,60} 
\definecolor{KITblack}{rgb}{0,0,0}
\definecolor{KITblue}{rgb}{.27,.39,.66}
\definecolor{KITred}{rgb}{.63,.13,.13}
\definecolor{KITpurple}{rgb}{.64,.06,.48}
\definecolor{KITcyan}{rgb}{.14,.63,.87}
\definecolor{KITyellow}{rgb}{.98,.89,0}
\definecolor{KITorange}{rgb}{.87,.60,.10}

\definecolor{Gold}{rgb}{1.0, 0.84, 0.0}
\definecolor{Periwinkle}{rgb}{0.6, 0.6, 1.0}
\definecolor{Orange}{rgb}{1.0, 0.65, 0.0}
\definecolor{Red}{rgb}{0.8, 0.1, 0.1}
\definecolor{Blue}{rgb}{0.2, 0.3, 0.8}

\colorlet{FFTBP_flooding}{KITpurple}
\colorlet{FFTBP_layered}{KITgreen}
\colorlet{EPDFT}{KITgreen}
\colorlet{ZF}{Blue}
\colorlet{MMSE_GaussPrior}{Orange}
\colorlet{MMSE_noPrior}{Red}
\colorlet{MAP}{black}

\newcommand\bfnlu{150} %
\newcommand\bfnru{30}
\newcommand\bfnlb{210}
\newcommand\bfnrb{-30}
\newsavebox\bfn %
\sbox\bfn{\begin{tikzpicture}
        \node[draw, fill=FFTBP_flooding!40, very thick, regular polygon, regular polygon sides=4, minimum width = 4em, inner sep=0pt, rounded corners](fn){};
        \draw[KITblack!70, shorten <= 0.12em, shorten >= 0.12em] (fn.\bfnlu) -- (fn.\bfnru) node (lu) [anchor=center,pos=0.1]{}  node (ru) [anchor=center,pos=0.9]{};
        \draw[KITblack!70, shorten <= 0.12em, shorten >= 0.12em] (fn.\bfnlb) -- (fn.\bfnrb) node (lb) [anchor=center,pos=0.1]{}  node (rb) [anchor=center,pos=0.9]{};
        \node[inner sep=0pt, minimum size = 1.3em, below=0em of fn.center, anchor=center](label){};
        \draw[KITblack!70] (lu.center) -- (label);
        \draw[KITblack!70] (ru.center) -- (label);
        \draw[KITblack!70] (lb.center) -- (label);
        \draw[KITblack!70] (rb.center) -- (label);
\end{tikzpicture}}
\newsavebox\largebfn %
\sbox\largebfn{\begin{tikzpicture}
        \node[draw, fill=FFTBP_flooding!40, very thick, regular polygon, regular polygon sides=4, minimum width = 9em, inner sep=0pt, rounded corners](fn){};
        \draw[KITblack!70, thick, shorten <= 0.12em, shorten >= 0.12em] (fn.\bfnlu) -- (fn.\bfnru) node (lu) [anchor=center,pos=0.1]{}  node (ru) [anchor=center,pos=0.9]{};
        \draw[KITblack!70, thick, shorten <= 0.12em, shorten >= 0.12em] (fn.\bfnlb) -- (fn.\bfnrb) node (lb) [anchor=center,pos=0.1]{}  node (rb) [anchor=center,pos=0.9]{};
        \node[inner sep=0pt, minimum size = 1.3em, below=0em of fn.center, anchor=center](label){};
        \draw[KITblack!70, thick] (lu.center) -- (label);
        \draw[KITblack!70, thick] (ru.center) -- (label);
        \draw[KITblack!70, thick] (lb.center) -- (label);
        \draw[KITblack!70, thick] (rb.center) -- (label);
\end{tikzpicture}}
\acrodef{APP}[APP]{a posteriori probability}
\acrodef{BP}[BP]{belief propagation}
\acrodef{BPSK}[BPSK]{binary phase-shift keying}
\acrodef{BF}[BF]{butterfly function}
\acrodef{DFT}[DFT]{discrete Fourier transform}
\acrodef{EP}[EP]{expectation propagation}
\acrodef{FFT}[FFT]{fast Fourier transform}
\acrodef{GaBP}[GaBP]{Gaussian BP}
\acrodef{GM}[GM]{Gaussian mixture}
\acrodef{ISI}[ISI]{inter-symbol interference}
\acrodef{KL}[KL]{Kullback-Leibler}
\acrodef{LLR}[LLR]{log-likelihood ratio}
\acrodef{MAP}[MAP]{maximum a posteriori}
\acrodef{LMMSE}[LMMSE]{linear minimum mean squared error}
\acrodef{MSE}[MSE]{mean squared error}
\acrodef{OFDM}[OFDM]{orthogonal frequency-division multiplexing}
\acrodef{SNR}[SNR]{signal-to-noise ratio}
\acrodef{SER}[SER]{symbol error rate}
\acrodef{ZF}[ZF]{zero-forcing}

\begin{document}

\title{Uncertainty\hspace{0.2em}Propagation\hspace{0.2em}in\hspace{0.2em}the\hspace{0.2em}Fast\hspace{0.19em}Fourier\hspace{0.19em}Transform \\
\thanks{This work has received funding in part from the European Research Council (ERC) under the European Union’s Horizon 2020 research and innovation programme (grant agreement No. 101001899) and in part from the German Federal Ministry of Education and Research (BMBF) within the project Open6GHub (grant agreement 16KISK010).}
}

\author{\IEEEauthorblockN{Luca Schmid, Charlotte Muth, and Laurent Schmalen}
\IEEEauthorblockA{Communications Engineering Lab, Karlsruhe Institute of Technology (KIT)\\ Hertzstr. 16, 76187 Karlsruhe, Germany, Email: \texttt{first.last@kit.edu}}
}

\newcommand{\TODO}[1]{\textcolor{blue}{#1}}
\newcommand{\openquestion}[1]{\textcolor{red}{#1}}
\definecolor{green}{RGB}{120,170,55} 
\newcommand{\idea}[1]{\textcolor{green}{#1}}
\newcommand{\streichbar}[1]{\textcolor{black}{#1}}

\maketitle

\begin{abstract}
We address the problem of uncertainty propagation in the discrete Fourier transform by modeling the fast Fourier transform as a factor graph. Building on this representation, we propose an efficient framework for approximate Bayesian inference using belief propagation (BP) and expectation propagation, extending its applicability beyond Gaussian assumptions. By leveraging an appropriate BP message representation and a suitable schedule, our method achieves stable convergence with accurate mean and variance estimates. Numerical experiments in representative scenarios from communications demonstrate the practical potential of the proposed framework for uncertainty-aware inference in probabilistic systems operating across both time and frequency domain.
\end{abstract}

\begin{IEEEkeywords}
Fast Fourier transform, factor graphs, Gaussian belief propagation, expectation propagation.
\end{IEEEkeywords}

\section{Introduction}
Frequency-domain analysis is fundamental in communications and signal processing, serving as the natural dual to time-domain processing.
In this context, the \ac{DFT} is widely used to switch to the frequency domain, in which many discrete systems can be represented more efficiently~\cite{proakis_digital_2007}.
Nevertheless, \emph{uncertainty propagation} in the \ac{DFT} remains surprisingly unexplored, with most approaches relying on coarse approximations 
or restricting inference to a single domain, which often increases complexity. 
For instance, factor graphs are employed in~\cite{worthen_unified_2001} to derive message-passing algorithms for channel estimation and decoding, but the time-domain representation becomes complex, with many cycles in the graph, even for basic communication channels.

An exception is \cite{storkey_generalised_nodate}, which studies missing data recovery in time sequences by incorporating prior knowledge of power spectra.
By formulating the \ac{FFT} as a Bayesian network and applying \ac{BP}, \cite{storkey_generalised_nodate} proposes an elegant method for exchanging uncertainty between the time and frequency domain.
Beyond its original application, the method has been used for a sparse representation of Reed-Solomon codes~\cite{yedidia_sparse_2004} and for the number-theoretic transform in cryptography~\cite{julius_side-channel_2024}. However, it has not yet gained widespread adoption in signal processing.
Two key limitations hinder its broader use. First, the method is restricted to Gaussian distributions, restricting its applicability. Second, the \ac{FFT} factor graph contains many short cycles, degrading \ac{BP} convergence and accuracy~\cite{kschischang_factor_2001}. To address the latter, \cite{storkey_generalised_nodate} proposed the application of generalized \ac{BP}\cite{yedidia_constructing_2005} to improve the convergence behavior. However, 
the algorithm still requires fine-tuning, e.g., \cite{storkey_generalised_nodate} adds jitter noise to the network or \cite{barber_stable_2007} proposes to use the auxiliary variable trick for better convergence.

Probabilistic inference has great potential in communications and signal processing. Building on~\cite{storkey_generalised_nodate}, we establish a framework for uncertainty propagation in the \ac{FFT}, extending it to non-Gaussian scenarios via the \ac{EP} method. 
For a more structured formulation of the \ac{GaBP} in the probabilistic \ac{FFT} setting, we use Forney-style factor graphs~\cite{loeliger_introduction_2004}, which support hierarchical modeling.
Furthermore, we use more general \ac{GaBP} messages than those used in~\cite{storkey_generalised_nodate} to better capture covariances between real and imaginary components, preserving uncertainty information that would otherwise be lost.
In addition, we explore an alternative \ac{BP} message-passing schedule, which leads to faster convergence and higher accuracy compared to the layered schedule in~\cite{storkey_generalised_nodate}. Our numerical simulations demonstrate that in inherently noisy settings---a common scenario in communications and signal processing---the proposed \ac{GaBP} algorithm remains numerically stable and yields accurate results. Remarkably, also the stochastic uncertainty in the form of covariances, often expected to be overconfident in loopy \ac{GaBP}, is very reliable.
Finally, we address open questions from~\cite{storkey_generalised_nodate}, exploring non-circular Gaussian priors and analyzing how the number of \ac{GaBP} iterations scales with the network size.

With this work, we aim to promote uncertainty propagation in the \ac{FFT}, unlocking its potential for various applications in communications and signal processing.

\section{Preliminaries} \label{sec:preliminaries}
We consider Bayesian inference in a probabilistic system ${p(\bm{\xt} | \bm{y}) \propto p(\bm{y} | \bm{\xt}) p(\bm{\xt})}$  with the latent variable of interest ${\bm{\xt}\in\mathbb{C}^N}$ and the observation~${\bm{y}}$.
We further assume that the likelihood function~$p(\bm{y} | \bm{\xt})$ can be expressed more conveniently (e.g., it factorizes) with respect to the transformed variable ${\bm{\xf} = \bm{W} \bm{\xt}}$, where $\WDFT$ is the symmetric $N$-point \ac{DFT} matrix~\cite{cormen2022introduction}. We call $\bm{\xt}$ the latent variable in \emph{time domain} and $\bm{\xf}$ the equivalent variable in \emph{frequency domain}.
With this, we can rewrite the \ac{APP} distribution
\begin{align}
    p(\bm{\xt} | \bm{y}) 
    &\propto \int p(\bm{y} | \bm{\xf}) \cdot \delta \mleft( \bm{\xf} - \bm{W}\bm{\xt} \mright) \cdot p(\bm{\xt}) \; \textrm{d}\bm{\xf}, \label{eq:likelihood_in_freq}
\end{align}
where we use the $\delta$-distribution to specify which pairs of the variables $\bm{\xt}$ and $\bm{\xf}$ are valid. 

In the following, we assume that inference over ${p(\bm{y} | \bm{\xf})}$ and ${p(\bm{\xt})}$ is tractable when considered separately\footnote{\streichbar{The alternative scenario where the prior and the likelihood function change domains, respectively, can be treated equivalently.}}.
Despite this assumption, inference over the \emph{full} \ac{APP} distribution~\eqref{eq:likelihood_in_freq} is generally still intractable or prohibitively complex for large~$N$. For instance, the \ac{DFT} imposes a combinatorial complexity for discrete probability mass functions.

\subsection{Uncertainty Propagation with Gaussian Distributions} \label{subsec:Gaussian_DFT_Uncertainty_propagation}
An exception is the multivariate Gaussian distribution, which is closed under any linear transformation, such as the \ac{DFT}~\cite{loeliger_factor_2007}. If the likelihood ${p(\bm{y}|\bm{\xf}) = \mathcal{N}(\bm{\xfreal}:\bm{\mu}_\textrm{f}, \bm{\Sigma}_\textrm{f})}$ is Gaussian with respect to ${\bm{\xf}}$, we obtain
\begin{equation*}
    p(\bm{y}|\bm{\xf}) \cdot \delta \mleft( \bm{\xf} \! - \! \bm{W}\bm{\xt} \mright)
    = \mathcal{N}\mleft(\bm{\xtreal}:\WIDFTreal \bm{\mu}_\textrm{f}, \WIDFTreal \bm{\Sigma}_\textrm{f} (\WIDFTreal)^\textrm{T}\mright),
\end{equation*}
where an underline represents the equivalent interleaved real-valued representation of a complex-valued vector or matrix
 \begin{align*}
 \underline{\bm{a}} &= \left(
         \textrm{Re}\mleft(a_1\mright), 
         \textrm{Im}\mleft(a_1\mright), 
         \textrm{Re}\mleft(a_2\mright), 
         \textrm{Im}\mleft(a_2\mright), 
         \ldots \right)^\textrm{T}, \\
     \underline{\bm{A}} &=  
     \begin{pmatrix}
     \tilde{\bm{A}}_{ij}
     \end{pmatrix}_{i,j}, 
     \quad
      \tilde{\bm{A}}_{ij} := \begin{pmatrix}
         \textrm{Re}\mleft({A}_{ij}\mright) & -\textrm{Im}\mleft({A}_{ij}\mright)\\
         \textrm{Im}\mleft({A}_{ij}\mright) & \textrm{Re}\mleft({A}_{ij}\mright)
     \end{pmatrix}.
 \end{align*}
If ${p(\bm{\xt}) = \mathcal{N}(\bm{\xtreal}:\bm{\mu}_\textrm{t}, \bm{\Sigma}_\textrm{t})}$ is also Gaussian, the full \ac{APP} distribution~\eqref{eq:likelihood_in_freq} can be expressed in closed form
\begin{align}
    p(\bm{\xt} | \bm{y}) 
    &= \mathcal{N}\mleft(\bm{\xtreal}: \bm{\mu}_\textrm{APP}, \bm{\Sigma}_\textrm{APP}\mright), \label{eq:Gaussian_closedForm} %
\end{align}
with the moments\footnote{It is numerically more stable to reformulate~\eqref{eq:mean_mult_of_2_gaussians} as the solution of a linear system of equations, rather than directly using the matrix inverse $\bm{\Sigma}_\textrm{APP}$~\cite{schmid2024fastrobustexpectationpropagation}. We adopt this approach for numerical stability but retain the notation as in~\eqref{eq:mean_mult_of_2_gaussians} involving the matrix inverse to avoid cluttering the presentation.}
\begin{align}
    \bm{\mu}_\textrm{APP} &= \bm{\Sigma}_\textrm{APP} \left( \WDFTreal^\textrm{T} \bm{\Sigma}^{-1}_\textrm{f} \bm{\mu}_\textrm{f} + \bm{\Sigma}_\textrm{t}^{-1}\bm{\mu}_\textrm{t} \right), \label{eq:mean_mult_of_2_gaussians} \\
    \bm{\Sigma}_\textrm{APP} &= \left( \WDFTreal^\textrm{T} \bm{\Sigma}^{-1}_\textrm{f} \WDFTreal + \bm{\Sigma}_\textrm{t}^{-1} \right)^{-1}. \label{eq:covariance_mult_of_2_gaussians} 
\end{align}
Based on~\eqref{eq:Gaussian_closedForm}, exact inference of the mean, \ac{MAP} inference ${\bm{\xt}_\textrm{MAP} =  \arg\max_{\bm{\xt}} p(\bm{\xt}|\bm{y})}$ and marginal inference are straightforward~\cite{bishop_pattern_2006}. 
The computational complexity scales with ${\mathcal{O}(N^3)}$ due to the matrix inversion in~\eqref{eq:covariance_mult_of_2_gaussians}.

\section{Expectation Propagation in the DFT}
If $p(\bm{y} | \bm{\xf})$ or $p(\bm{\xt})$ are non-Gaussian, we can approximate the marginals of the system's latent variables $\bm{\xt}$ and $\bm{\xf}$ by one-dimensional Gaussians,
introducing the global approximation
\begin{equation}
    p(\bm{\xt} | \bm{y}) \approx q(\bm{\xt}) \propto \prod\limits_{n=1}^N q(\xt_n) \cdot \delta \mleft( \bm{\xf} \! - \! \bm{W}\bm{\xt} \mright) \cdot \prod\limits_{n=1}^N q(\xf_n), \label{eq:global_Gaussian_approx}
\end{equation}
where the local factors 
${q(\xt_n) = \mathcal{N}_\textrm{can}\mleft(\underline{u}^\textrm{t}_n: \bm{\gamma}_n, \bm{\lambda}_n \mright)}$ and ${q(\xf_n) = \mathcal{N}_\textrm{can}\mleft(\underline{u}_n^\textrm{f}: \bm{\Gamma}_n, \bm{\Lambda}_n \mright)}$
are complex Gaussians 
\begin{equation*}
     \mathcal{N}_\textrm{can}(\underline{u}^\textrm{t}: \bm{\gamma}, \bm{\lambda}) := \exp \mleft(a + \bm{\gamma}^\textrm{T} \underline{u}^\textrm{t} -\frac{1}{2} {\underline{u}^\textrm{t}}^\textrm{T} \bm{\lambda} \underline{u}^\textrm{t} \mright),
\end{equation*} 
in canonical form with normalizing constant~$a$. %
We aim to find the information vectors ${\bm{\gamma}_n, \bm{\Gamma}_n \in \mathbb{R}^2}$ and the positive definite precision matrices ${\bm{\lambda}_n, \bm{\Lambda}_n \in \mathbb{R}^{2 \times 2} \succ 0}$
such that the global approximation $q(\bm{\xt})$ minimizes the \ac{KL} divergence ${D_\textrm{KL}\mleft( p(\bm{\xt} | \bm{y}) \Vert q(\bm{\xt}) \mright)}$~\cite{bishop_pattern_2006}.
The \ac{EP} method~\cite{minka_expectation_2013} approaches this task by iteratively refining the local parameters of the approximating factors $q(\xt_n)$ and $ q(\xf_n)$, until their moments are consistent throughout the global system.
\begin{algorithm}[t]
    \DontPrintSemicolon
    \KwData{Local distribution $p(x), \; x \in \mathbb{C}$,\\
    marginals of global approximation $\mathcal{N}\mleft(\underline{x}: \bm{\mu},\bm{\Sigma}  \mright)$, \\
    old EP parameters $\bm{\gamma}^\textrm{old}, \bm{\lambda}^\textrm{old}$}
    \vspace{3pt} 
        $\bm{\gamma}^\textrm{cav} = \bm{\Sigma}^{-1} \bm{\mu} - \bm{\gamma}^\textrm{old}, \; \bm{\lambda}^\textrm{cav} = \bm{\Sigma}^{-1} - \bm{\lambda}^\textrm{old}$ \\ \vspace{1pt}
        Compute the mean~$\bm{\hat{\mu}}$ and the covariance~$\bm{\hat{\Sigma}}$ of the distribution  $\hat{p}(x) \propto p(x) \cdot \mathcal{N}_\textrm{can}\mleft(\underline{x}: \bm{\gamma}^\textrm{cav}, \bm{\lambda}^\textrm{cav} \mright)$. \\ \vspace{1pt}
        $\bm{\gamma} = \bm{\hat{\Sigma}}^{-1} \bm{\hat{\mu}} - \bm{\gamma}^\textrm{cav} , \; \bm{\lambda} = \bm{\hat{\Sigma}}^{-1} - \bm{\lambda}^\textrm{cav}$\\  \vspace{1pt}
        \lIf{$\bm{\lambda} \not\succ 0$}
        {
           $\, \bm{\lambda} = \bm{\lambda}^\textrm{old}, \; \bm{\lambda} = \bm{\lambda}^\textrm{old}$
        } \vspace{1pt}
        $\bm{\lambda}^\textrm{new} =  \beta \bm{\lambda} + (1-\beta) \bm{\lambda}^\textrm{old}, \; \bm{\gamma}^\textrm{new} =  \beta \bm{\gamma} + (1-\beta) \bm{\gamma}^\textrm{old}$ \\ \vspace{3pt}
    \KwResult{Updated EP parameters $\bm{\gamma}^\textrm{new}, \bm{\lambda}^\textrm{new}$}
    \caption{EP} 
    \label{alg:EP_update}
\end{algorithm}

Alg.~\ref{alg:EP_update} defines one parameter update of the iterative \ac{EP} method in our setting. 
Given the marginals of the current global approximation, the algorithm updates the local parameters ${\bm{\gamma},\bm{\lambda}}$ such that the moments of the new approximation ${\mathcal{N}_\textrm{can}\mleft(\underline{x}: \bm{\gamma}^\textrm{cav}, \bm{\lambda}^\textrm{cav} \mright) \cdot \mathcal{N}_\textrm{can}\mleft(\underline{x}: \bm{\gamma}, \bm{\lambda} \mright)}$ match with the moments of the distribution ${\mathcal{N}_\textrm{can}\mleft(\underline{x}: \bm{\gamma}^\textrm{cav}, \bm{\lambda}^\textrm{cav} \mright) \cdot p(x)}$, which incorporates the true local distribution.
Finally, in lines~4-5 of Alg.~\ref{alg:EP_update}, we check the validity of the new precision matrix and smooth the parameter updates with ${\beta=0.5}$ to improve convergence. 

Combining the \ac{EP} updates in Alg.~\ref{alg:EP_update} and the Gaussian uncertainty propagation in the \ac{DFT} in Sec.~\ref{subsec:Gaussian_DFT_Uncertainty_propagation}, we define the EP-DFT algorithm:
after initializing the parameters ${\bm{\gamma}_n = \bm{\Gamma}_n = \bm{0}_{2}}$, ${\bm{\lambda}_n = \bm{\Lambda}_n = \bm{0}_{2,2}}$ for all local Gaussian approximations ${q(\xt_n)}$ and ${q(\xf_n)}$, the EP-DFT algorithm iteratively alternates between simultaneously refining all parameters using Alg.~\ref{alg:EP_update}, and putting the local approximations into a global context, by constructing the global Gaussian approximation~\eqref{eq:global_Gaussian_approx} in both time and frequency domain.

\section{Belief Propagation on the FFT Factor Graph}
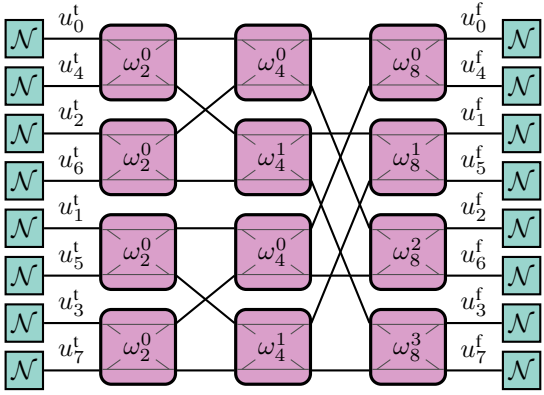
\begin{figure}[tb]
\centering
\tikzstyle{butterfly_fn} = [draw, very thick, regular polygon, regular polygon sides=4, minimum width = 4em, inner sep=0pt, rounded corners]
\tikzstyle{vn} = [inner sep=0pt, minimum size = 1em]

\tikzstyle{vn} = [draw, fill=KITgreen!40, thick, regular polygon, regular polygon sides=4, minimum width = 2em, inner sep=0pt]

\begin{tikzpicture}[auto, node distance=0.5em and 2em, thick]

    \node[inner sep=0pt](b11){\usebox{\bfn}};
    \node[inner sep=0pt, below=0.56em of b11](b12){\usebox{\bfn}}; 
    \node[inner sep=0pt, below=0.56em of b12](b13){\usebox{\bfn}};
    \node[inner sep=0pt, below=0.56em of b13](b14){\usebox{\bfn}};
    \node[below=0em of b11.center, anchor=center](){$\omega_2^0$};
    \node[below=0em of b12.center, anchor=center](){$\omega_2^0$};
    \node[below=0em of b13.center, anchor=center](){$\omega_2^0$};
    \node[below=0em of b14.center, anchor=center](){$\omega_2^0$};           
    \node [vn, left=of b11.\bfnlu] (f0){$\mathcal{N}$};
    \node [vn, left=of b11.\bfnlb] (f1){$\mathcal{N}$};
    \node [vn, left=of b12.\bfnlu] (f2){$\mathcal{N}$};
    \node [vn, left=of b12.\bfnlb] (f3){$\mathcal{N}$};
    \node [vn, left=of b13.\bfnlu] (f4){$\mathcal{N}$};
    \node [vn, left=of b13.\bfnlb] (f5){$\mathcal{N}$};
    \node [vn, left=of b14.\bfnlu] (f6){$\mathcal{N}$};
    \node [vn, left=of b14.\bfnlb] (f7){$\mathcal{N}$};
    \draw (f0.east) -- node[above=-1pt]{$\xt_0$} ([xshift=0.1em]b11.\bfnlu);
    \draw (f1.east) -- node[above=-1pt]{$\xt_4$} ([xshift=0.1em]b11.\bfnlb);
    \draw (f2.east) -- node[above=-1pt]{$\xt_2$} ([xshift=0.1em]b12.\bfnlu);
    \draw (f3.east) -- node[above=-1pt]{$\xt_6$} ([xshift=0.1em]b12.\bfnlb);
    \draw (f4.east) -- node[above=-1pt]{$\xt_1$} ([xshift=0.1em]b13.\bfnlu);
    \draw (f5.east) -- node[above=-1pt]{$\xt_5$} ([xshift=0.1em]b13.\bfnlb);
    \draw (f6.east) -- node[above=-1pt]{$\xt_3$} ([xshift=0.1em]b14.\bfnlu);
    \draw (f7.east) -- node[above=-1pt]{$\xt_7$} ([xshift=0.1em]b14.\bfnlb);
    \node [inner sep=0pt, right=of b11](b21){\usebox{\bfn}};
    \node [inner sep=0pt, right=of b12](b22){\usebox{\bfn}};
    \node [inner sep=0pt, right=of b13](b23){\usebox{\bfn}};
    \node [inner sep=0pt, right=of b14](b24){\usebox{\bfn}};
    \node[below=0em of b21.center, anchor=center](){$\omega_4^0$};
    \node[below=0em of b22.center, anchor=center](){$\omega_4^1$};
    \node[below=0em of b23.center, anchor=center](){$\omega_4^0$};
    \node[below=0em of b24.center, anchor=center](){$\omega_4^1$};
    \draw ([xshift=-0.1em]b11.\bfnru) -- ([xshift=0.1em]b21.\bfnlu);
    \draw ([xshift=-0.1em]b11.\bfnrb) -- ([xshift=0.1em]b22.\bfnlu);
    \draw ([xshift=-0.1em]b12.\bfnru) -- ([xshift=0.1em]b21.\bfnlb);
    \draw ([xshift=-0.1em]b12.\bfnrb) -- ([xshift=0.1em]b22.\bfnlb);
    \draw ([xshift=-0.1em]b13.\bfnru) -- ([xshift=0.1em]b23.\bfnlu);
    \draw ([xshift=-0.1em]b13.\bfnrb) -- ([xshift=0.1em]b24.\bfnlu);
    \draw ([xshift=-0.1em]b14.\bfnru) -- ([xshift=0.1em]b23.\bfnlb);
    \draw ([xshift=-0.1em]b14.\bfnrb) -- ([xshift=0.1em]b24.\bfnlb);
    \node [inner sep=0pt, right=of b21](b31){\usebox{\bfn}};
    \node [inner sep=0pt, right=of b22](b32){\usebox{\bfn}};
    \node [inner sep=0pt, right=of b23](b33){\usebox{\bfn}};
    \node [inner sep=0pt, right=of b24](b34){\usebox{\bfn}};
    \node[below=0em of b31.center, anchor=center](){$\omega_8^0$};
    \node[below=0em of b32.center, anchor=center](){$\omega_8^1$};
    \node[below=0em of b33.center, anchor=center](){$\omega_8^2$};
    \node[below=0em of b34.center, anchor=center](){$\omega_8^3$};
    \draw ([xshift=-0.1em]b21.\bfnru) -- ([xshift=0.1em]b31.\bfnlu);
    \draw ([xshift=-0.1em]b21.\bfnrb) -- ([xshift=0.1em]b33.\bfnlu);
    \draw ([xshift=-0.1em]b22.\bfnru) -- ([xshift=0.1em]b32.\bfnlu);
    \draw ([xshift=-0.1em]b22.\bfnrb) -- ([xshift=0.1em]b34.\bfnlu);
    \draw ([xshift=-0.1em]b23.\bfnru) -- ([xshift=0.1em]b31.\bfnlb);
    \draw ([xshift=-0.1em]b23.\bfnrb) -- ([xshift=0.1em]b33.\bfnlb);
    \draw ([xshift=-0.1em]b24.\bfnru) -- ([xshift=0.1em]b32.\bfnlb);
    \draw ([xshift=-0.1em]b24.\bfnrb) -- ([xshift=0.1em]b34.\bfnlb);
    \node [vn, right=of b31.\bfnru] (F0){$\mathcal{N}$};
    \node [vn, right=of b31.\bfnrb] (F1){$\mathcal{N}$};
    \node [vn, right=of b32.\bfnru] (F2){$\mathcal{N}$};
    \node [vn, right=of b32.\bfnrb] (F3){$\mathcal{N}$};
    \node [vn, right=of b33.\bfnru] (F4){$\mathcal{N}$};
    \node [vn, right=of b33.\bfnrb] (F5){$\mathcal{N}$};
    \node [vn, right=of b34.\bfnru] (F6){$\mathcal{N}$};
    \node [vn, right=of b34.\bfnrb] (F7){$\mathcal{N}$};  
    \draw ([xshift=-0.1em]b31.\bfnru) -- node[above=-1pt]{$\xf_0$} (F0.west);
    \draw ([xshift=-0.1em]b31.\bfnrb) -- node[above=-1pt]{$\xf_4$} (F1.west);
    \draw ([xshift=-0.1em]b32.\bfnru) -- node[above=-1pt]{$\xf_1$} (F2.west);
    \draw ([xshift=-0.1em]b32.\bfnrb) -- node[above=-1pt]{$\xf_5$} (F3.west);
    \draw ([xshift=-0.1em]b33.\bfnru) -- node[above=-1pt]{$\xf_2$} (F4.west);
    \draw ([xshift=-0.1em]b33.\bfnrb) -- node[above=-1pt]{$\xf_6$} (F5.west);
    \draw ([xshift=-0.1em]b34.\bfnru) -- node[above=-1pt]{$\xf_3$} (F6.west);
    \draw ([xshift=-0.1em]b34.\bfnrb) -- node[above=-1pt]{$\xf_7$} (F7.west);
\end{tikzpicture}
\caption{Forney-style factor graph of~\eqref{eq:global_Gaussian_approx} for ${N=8}$ using the \ac{FFT}.}
    \label{fig:fft_factor_graph}
\end{figure}

\begin{figure}[tb]
\setlength{\abovecaptionskip}{-2pt}
\centering
\tikzstyle{butterfly_fn} = [draw, very thick, dashed, regular polygon, regular polygon sides=4, minimum width = 4em, inner sep=0pt]
\tikzstyle{vn} = [inner sep=0pt, minimum size = 0em]
\begin{tikzpicture}
        \node[draw, fill=FFTBP_flooding!40, regular polygon, regular polygon sides=4, minimum width = 9em, inner sep=0pt, dashed](fn){};
        \node[vn, left=2em of fn.\bfnlu](x0){};
        \node[vn, right=2em of fn.\bfnru](y0){};
        \draw[thick] (x0) -- node[above=-1pt, pos=0.08]{$x_0$} node[above=-1pt, pos=0.92]{$y_0$} (y0);
        \node[draw, fill=gray!20, thick, regular polygon, regular polygon sides=4, minimum width = 2em, inner sep=0pt, right=1em of fn.\bfnlu, anchor=west](eu){$=$};
        \node[draw, fill=gray!20, thick, regular polygon, regular polygon sides=4, minimum width = 2em, inner sep=0pt, left=1em of fn.\bfnru, anchor=east](mu){$+$};
        \node[vn, left=2em of fn.\bfnlb](x1){};
        \node[right=2em of fn.\bfnrb](y1){};
        \draw[thick] (x1) -- node[above=-1pt, pos=0.08]{$x_1$} node[above=-1pt, pos=0.92]{$y_1$} (y1);
        \node[draw, fill=gray!20, thick, regular polygon, regular polygon sides=4, minimum width = 2em, inner sep=0pt, right=1em of fn.\bfnlb, anchor=west](eb){$=$};
        \node[draw, fill=gray!20, thick, regular polygon, regular polygon sides=4, minimum width = 2em, inner sep=0pt, left=1em of fn.\bfnrb, anchor=east](mb){$-$};
        \node[draw, fill=gray!20, thick, regular polygon, regular polygon sides=4, minimum width = 2em, inner sep=0pt, right=1em of fn.west, anchor=west](wl){$\omega$};
        \node[draw, fill=gray!20, thick, regular polygon, regular polygon sides=4, minimum width = 2em, inner sep=0pt, left=1em of fn.east, anchor=east](wr){$\omega$};
        \draw[thick] (eu) -- (wl) -- (mb);
        \draw[thick] (eb) -- (wr) -- (mu);
\node[vn, right=2em of fn, anchor=center](arrow){};
\node[right = 7em of fn, inner sep=0pt](b11){\usebox{\largebfn}};
\node[below=0em of b11.center, anchor=center](){$\omega_n^k$};
\node [vn, left=2.5em of b11.\bfnlu] (x0){};
\node [vn, left=2.5em of b11.\bfnlb] (x1){};
\node [vn, right=2.5em of b11.\bfnru] (y00){};
\node [vn, right=2.5em of b11.\bfnrb] (y1){};
\draw[thick] ([xshift=0.2em]x0.east) --  ([xshift=0.1em]b11.\bfnlu);
\draw[-latex, thick, KITgreen] ([xshift=0.5em, yshift=0.4em]x0.east) -- node[above, font=\small] {$\tiny\nu(x_0)$} ([xshift=-0.5em, yshift=0.4em]b11.\bfnlu);
\draw[-latex, thick, KITpurple] ([xshift=-0.5em, yshift=-0.4em]b11.\bfnlu) -- node[below, font=\small] {$\xi(x_0)$} ([xshift=0.5em, yshift=-0.4em]x0.east);
\draw[thick] ([xshift=0.5em]x1.east) -- ([xshift=0.1em]b11.\bfnlb);
\draw[-latex, thick, KITgreen] ([xshift=0.5em, yshift=0.4em]x1.east) -- node[above, font=\small] {$\nu(x_1)$} ([xshift=-0.5em, yshift=0.4em]b11.\bfnlb);
\draw[-latex, thick, KITpurple] ([xshift=-0.5em, yshift=-0.4em]b11.\bfnlb) -- node[below, font=\small] {$\xi(x_1)$} ([xshift=0.5em, yshift=-0.4em]x1.east);
\draw[thick] ([xshift=-0.5em]y00.west) -- ([xshift=-0.1em]b11.\bfnru);
\draw[-latex, thick, KITgreen] ([xshift=-0.5em, yshift=0.4em]y00.west) -- node[above, font=\small] {$\nu(y_0)$} ([xshift=0.5em, yshift=0.4em]b11.\bfnru);
\draw[-latex, thick, KITpurple] ([xshift=0.5em, yshift=-0.4em]b11.\bfnru) -- node[below, font=\small] {$\xi(y_0)$} ([xshift=-0.5em, yshift=-0.4em]y00.west);
\draw[thick] ([xshift=-0.5em]y1.west) -- ([xshift=-0.1em]b11.\bfnrb);
\draw[-latex, thick, KITgreen] ([xshift=-0.5em, yshift=0.4em]y1.west) -- node[above, font=\small] {$\nu(y_1)$} ([xshift=0.5em, yshift=0.4em]b11.\bfnrb);
\draw[-latex, thick, KITpurple] ([xshift=0.5em, yshift=-0.4em]b11.\bfnrb) -- node[below, font=\small] {$\xi(y_1)$} ([xshift=-0.5em, yshift=-0.4em]y1.west);
\draw[dashed, gray] ([yshift=-5.2em]$(y0)!0.5!(x0)$) -- ([yshift=1.7em]$(y0)!0.5!(x0)$);
\end{tikzpicture}
\caption{Left: hierarchical modeling (``boxes within boxes''~\cite{loeliger_introduction_2004}) of the \ac{BF} in~\eqref{eq:butterfly_factor}. Right: clustered \ac{BF} node with in- and outgoing \ac{BP} messages.}
    \label{fig:fft_butterfly_node}
\end{figure}
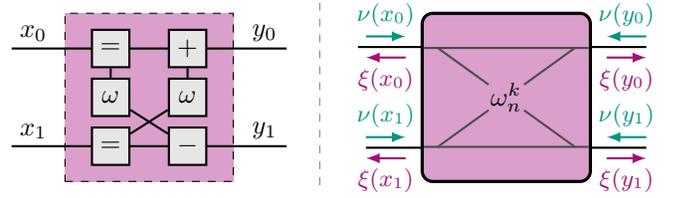

A more efficient way to compute the \ac{DFT} 
is provided by the ubiquitous \ac{FFT}~\cite{cormen2022introduction}.
Employing a divide-and-conquer strategy together with an efficient reordering of even- and odd-indexed subsequences, the \ac{FFT} breaks down the \ac{DFT} into basic, so-called \acp{BF}
\begin{equation}
    \begin{pmatrix} {y_0} \\ {y_1} \end{pmatrix} = 
    \underbrace{
    \begin{pmatrix} {1} & {\omega_n^k} \\ {1} & -{\omega_n^k} \end{pmatrix}}_{=: \bm{B}} %
    \begin{pmatrix} {x_0} \\ {x_1} \end{pmatrix}, \quad \omega_n^k := \exp \mleft( -\textrm{j} 2\pi k / n \mright).
    \label{eq:butterfly_factor}
\end{equation}
Using this decomposition, we can graphically represent the factorization in~\eqref{eq:global_Gaussian_approx} in a more fine-grained manner using a factor graph, as visualized in Fig.~\ref{fig:fft_factor_graph}.
We use a Forney-style factor graph~\cite{loeliger_introduction_2004}, as it naturally reproduces the graphical structure of the deterministic \ac{FFT} circuit~\cite{cormen2022introduction}.
Undirected edges represent variables and nodes represent the local functions of the underlying factorization. Incident edges to a node indicate on which variables the respective local function depends.
Forney-style factor graphs are well-suited for hierarchical modeling of the \ac{BF}~\eqref{eq:butterfly_factor}, as visualized in Fig.~\ref{fig:fft_butterfly_node}.
Applying \ac{BP} on the \emph{clustered} \ac{BF} nodes is motivated due to the short cycles within the subgraph of the \ac{BF} node, which otherwise causes poor convergence behavior of the \ac{BP} algorithm.
As noted in~\cite{storkey_generalised_nodate, yedidia_sparse_2004}, this can be considered a simple form of generalized \ac{BP}~\cite{yedidia_constructing_2005}.

We apply the \ac{BP} algorithm~\cite{kschischang_factor_2001} on the factor graph in Fig.~\ref{fig:fft_factor_graph}.
Since it only consists of Gaussian factors and linear \ac{BF} building blocks, the model can be interpreted as a general form of \emph{Kalman filtering}~\cite{loeliger_introduction_2004}, itself a specific instance of \ac{GaBP}\footnote{Technically, \ac{GaBP} is defined as \ac{BP} on a factor graph in which \emph{all} factors are Gaussian~\cite{weiss_correctness_2001}. However, the linear building blocks of the generalized Kalman filter can be viewed as limits of Gaussians, as outlined in~\cite{loeliger_least_2002}.}. To represent the \ac{GaBP} messages, \cite{storkey_generalised_nodate} uses \mbox{1-dimensional} complex Gaussians\footnote{The $M$-dimensional complex Gaussian is defined as ${(\pi^M \textrm{det}\bm{Z})^{-1} \textrm{exp}(-\bm{z}^\textrm{H} \bm{Z}^{-1} \bm{z}/2) }$ where ${\bm{z} \in \mathbb{C}^M}$ and ${\bm{Z} \in \mathbb{C}^{M \times M} \succcurlyeq 0}$.}, which are restrictive in capturing the covariances between real and imaginary parts of variabes~\cite{van_den_bos_multivariate_1995}. To avoid a loss of uncertainty information in every \ac{BP} update, we use the generalized representation by \mbox{2-dimensional} real-valued Gaussians~\cite{van_den_bos_multivariate_1995}.
We initialize the \ac{GaBP} algorithm by setting all \ac{BP} messages between the \ac{BF} nodes to ${\mathcal{N}\mleft( \underline{z}: \bm{0}_2, \infty \cdot \bm{I}_2 \mright)}$, where we replace $\infty$ with a large number in numerical implementations.
The messages leaving the \mbox{degree-1} Gaussian factor nodes at the boundaries of the graph are $q(\xt_n)$ and $q(\xf_n)$ and remain constant.

The \ac{BP} message updates for the introduced \ac{BF} nodes follow by applying the sum-product rule~\cite{loeliger_introduction_2004}.
As shown in Fig.~\ref{fig:fft_butterfly_node}, we denote incident messages by~${\nu(z) = \mathcal{N}\mleft(\underline{z}: \bm{\mu}_z, \bm{\Sigma}_z  \mright)}$ and outgoing messages by~${\xi(z)}$.
The outgoing message~${\xi(y_1)}$ is updated as a function of all extrinsic incoming messages: 
\begin{align}
    \xi(y_1) &= \! \iiint 
    \delta \mleft( \! \! \begin{pmatrix} {y_0} \\ {y_1} \end{pmatrix} - \bm{B} \begin{pmatrix} {x_0} \\ {x_1} \end{pmatrix} \! \! \mright)
    \prod\limits_{z \in \{y_0,x_0, x_1\}} \hspace{-15pt} \nu(z) \,
      \textrm{d}y_0 \, \textrm{d}x_0 \, \textrm{d}x_1 \nonumber \\
      &= \hspace{-0.4em} \int \hspace{-0.2em}
        \mathcal{N}\Bigg( \hspace{-0.6em} \begin{pmatrix} \underline{y}_0 \\ \underline{y}_1 \end{pmatrix} \hspace{-0.2em} :  
        \hspace{-0.1em} \underline{\bm{B}} \hspace{-0.1em}
        \begin{pmatrix} \bm{\mu}_{x_0} \\ \bm{\mu}_{x_1} \end{pmatrix} \hspace{-0.1em},
        \underbrace{\underline{\bm{B}} \hspace{-0.1em}
        \setlength{\arraycolsep}{0.05em} %
            \begin{pmatrix} 
            \bm{\Sigma}_{x_0} & \bm{0} \\
            \bm{0} & \bm{\Sigma}_{x_1}
            \end{pmatrix}
        \hspace{-0.1em} \underline{\bm{B}}^\textrm{T}}_{=: \bm{\Lambda}^{-1}_{y}}
        \hspace{-0.2em} \Bigg)
         \nu(y_0)
      \textrm{d}y_0 \nonumber \vspace{-5em} \\
      &= \mathcal{N} \Big( \underline{y}_1 : \bm{\mu}_{y[3:4]}, \bm{\Sigma}_{y[3:4],[3:4]} \Big), \label{eq:marginal_after_linear_trafo}
\end{align}
with mean and covariance
\begin{align*}
            \bm{\mu}_y &:= \bm{\Sigma}_y \mleft( \bm{\Lambda}_{y} 
            \underline{\bm{B}}
            \begin{pmatrix} \bm{\mu}_{x_0} \\ \bm{\mu}_{x_1} \end{pmatrix} 
            + \begin{pmatrix} \bm{\Sigma}_{y_0}^{-1} \bm{\mu}_{y_0} \\ \bm{0} \end{pmatrix}
            \mright),\\
     \bm{\Sigma}_y &:= \mleft( \bm{\Lambda}_{y} + 
            \begin{pmatrix} 
            \bm{\Sigma}_{y_0}^{-1} & \bm{0} \\
            \bm{0} & \bm{0}
            \end{pmatrix} \mright)^{-1}.
\end{align*}
The update rule for $\xi(y_0)$ follows from symmetry by replacing $\nu(y_0)$ with $\nu(y_1)$ in the product of the incoming messages and marginalizing out $y_1$ in~\eqref{eq:marginal_after_linear_trafo}: $\xi(y_0) = {\mathcal{N}\big( \underline{y}_0 : \bm{\mu}_{y[1:2]}, \bm{\Sigma}_{y[1:2],[1:2]} \big)}$. 
For updating~$\xi(x_0)$ and $\xi(x_1)$, the same steps as in~\eqref{eq:marginal_after_linear_trafo} apply, considering the inverse \ac{BF} transformation, i.e., all $x$ and $y$ dependencies are interchanged and $\bm{B}$ is replaced by $\bm{B}^{-1}$.

The \ac{BP} algorithm iteratively updates the messages in the factor graph until convergence\footnote{We declare a \ac{GaBP} message~${\mathcal{N}(\underline{z}: \bm{\mu}, \bm{\Sigma})}$ converged if all its scalar parameters~${m^{(t)} \in \{\mu_i, \Sigma_{ij}: i,j=1,2\}}$ in iteration $t$ are equivalent to the parameters $m^{(t-1)}$ in the previous iteration up to a precision tolerance $\tau_\textrm{conv} = 10^{-5}$, i.e., $\big| m^{(t)} - m^{(t-1)}\big| \leq \tau_\textrm{conv} \left(1 + \big| m^{(t-1)} \big|\right)$.}.
We consider two distinct message passing schedules: the \emph{flooding} schedule, where all messages in the graph are updated simultaneously~\cite{kschischang_factor_2001}, and the \emph{layered} schedule~\cite{storkey_generalised_nodate}, where messages propagate from left to right and then back. 
In the latter, updating all messages once requires ${(2 \log_2 N -1)}$ \ac{BP} iterations which we call one \emph{layered iteration}.

After convergence, the \ac{BP} algorithm computes the beliefs $b(\xt_n)$ and $b(\xf_n)$ by multiplying the two messages with opposing directions on the edges $\xt_n$ and $\xf_n$, respectively, followed by normalization.
In \ac{GaBP}, the sum-product and max-product algorithms coincide~\cite{loeliger_introduction_2004}, i.e., the beliefs represent both the single-variable marginals and the global maximizing configuration.
The factor graph in Fig.~\ref{fig:fft_factor_graph} contains cycles, so \ac{GaBP} is not guaranteed to converge~\cite{kschischang_factor_2001}. However, if it does, the means are exact~\cite{weiss_correctness_2001}. The covariances lack such guarantees and are typically assumed to be overconfident~\cite{weiss_correctness_2001, loeliger_introduction_2004}.

Leveraging the \ac{GaBP} algorithm on the \ac{FFT} factor graph for Gaussian uncertainty propagation in the EP-DFT scheme, we propose the EP-FFT algorithm as a low-complexity variant of the EP-DFT baseline in Alg.~\ref{alg:EP-FFT}.
We can achieve further efficiency gains by retaining the \ac{GaBP} messages between the runs in line~3 of Alg.~\ref{alg:EP-FFT},
reducing the required \ac{BP} iterations for convergence, especially in later \ac{EP} iterations. 
\begin{algorithm}[t]
    \DontPrintSemicolon
    \KwData{$p(\bm{y} | \bm{\xf})$, $p(\bm{\xt})$}\vspace{1pt}
    $\bm{\gamma}_n = \bm{\Gamma}_n = \bm{0}_{2}$, $\bm{\lambda}_n = \bm{\Lambda}_n = \bm{0}_{2,2}, \quad n=1,\ldots N$\\ \vspace{1pt} 
    \For{$\ell = 1,\ldots,L$} 
    {   \vspace{1pt}
        Run GaBP on the factor graph representation of~\eqref{eq:global_Gaussian_approx} as in Fig.~\ref{fig:fft_factor_graph}, to compute the marginal beliefs $b^{(\ell)}(\xt_n)$ and $b^{(\ell)}(\xf_n)$. \vspace{1pt}\\
        EP parameter updates: $\;$ for $n=1,\ldots,N$ \vspace{-15pt}\\
        \nonl \begin{align*}
            \bm{\gamma}_n, \bm{\lambda}_{n} &= \textrm{EP}\mleft( p(\xt_n), b^{(\ell)}(\xt_n), \bm{\gamma}_n, \bm{\lambda}_{n} \mright) \\
            \bm{\Gamma}_n, \bm{\Lambda}_{n} &= \textrm{EP}\mleft( p(\bm{y} | \xf_n), b^{(\ell)}(\xf_n), \bm{\Gamma}_n, \bm{\Lambda}_{n} \mright)
        \end{align*} \vspace{-15pt}
    }
    \KwResult{$b^{(L)}(\xt_n), \quad n=1,\ldots,N$} 
    \caption{EP-FFT} 
    \label{alg:EP-FFT}
\end{algorithm}

\section{Numerical Simulations}
We first analyze the convergence behavior and precision of \ac{GaBP} on the FFT factor graph. Then, we examine two exemplary scenarios from communications where the EP-FFT framework enables efficient approximate inference.
We provide the source code for all simulations in~\cite{github_2025}.

\subsection{Accuracy of GaBP in the FFT Factor Graph}
To evaluate \ac{GaBP} on the \ac{FFT} factor graph, we sample $100$ pairs of trial distributions ${p(\bm{\xt})=\prod_{n=1}^N \mathcal{N}\mleft(\underline{u}_n^\textrm{t}: \bm{\mu}_n^\textrm{t}, \bm{\Sigma}_n^\textrm{t}  \mright)}$ and ${p(\bm{\xf}) = \prod_{n=1}^N \mathcal{N}\mleft(\underline{u}^\textrm{f}_n: \bm{\mu}_n^\textrm{f}, \bm{\Sigma}_n^\textrm{f}  \mright)}$. 
The time-domain covariances~$\bm{\Sigma}_n^\textrm{t}$ are diagonal with elements uniformly sampled from ${[0,1]}$, while the frequency-domain covariances~$\bm{\Sigma}_n^\textrm{f}$ are sampled from ${[0,N]}$.
The means~$\bm{\mu}_n^\textrm{f}$ are drawn from~${\mathcal{N}(\bm{\mu}_n^\textrm{f}:\bm{0}_2,\bm{\Sigma}_n^\textrm{f})}$
, while the time-domain means are noisy versions of ${\WIDFTreal \bm{\mu}_n^\textrm{f}}$, with additive Gaussian noise sampled from a zero-mean Gaussian with covariance $\bm{\Sigma}_n^\textrm{f}$.

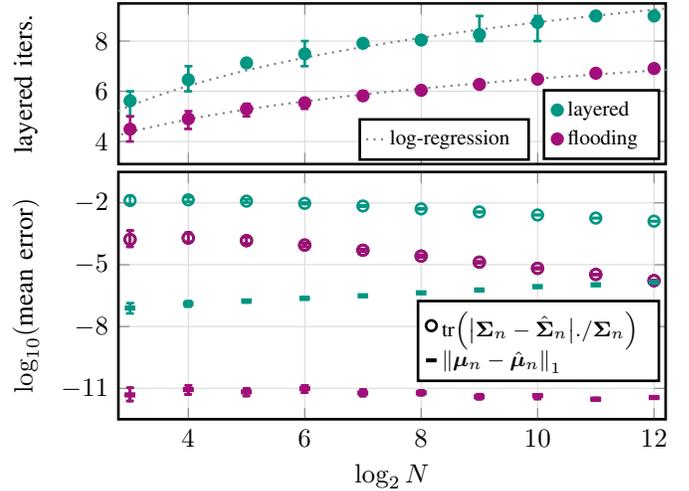
\begin{figure}[tb]
\setlength{\abovecaptionskip}{-3pt} %
\centering
  \begin{tikzpicture}
\begin{groupplot}[
        group style={
            group size=1 by 2,
            x descriptions at=edge bottom,
            vertical sep=3pt,
        },
    ] 
  \nextgroupplot[ %
    width=\linewidth, %
    height=0.42\linewidth,
    align = left,
    grid=major, %
    grid style={gray!30}, %
    ylabel= layered iters.,
    ymin = 3.1,
    ymax = 9.5,
    xmin = 2.8,
    xmax = 12.2,
    enlarge x limits=false,
    enlarge y limits=false,
    line width=1pt,
	  legend style={font=\footnotesize, cells={align=left}, anchor=south east, at={(0.99,0.025)}},
    legend cell align={left},
    ]
    \addplot[FFTBP_layered,only marks, error bars/.cd, y dir=both, y explicit, error bar style={line width=1pt}, error mark options={rotate=90,FFTBP_layered,mark size=1.5pt, line width=1pt}] table[x={log2N}, y={mean_iters_layered}, y error plus={q75_iters_layered}, y error minus={q25_iters_layered}, col sep=comma] {numerical_results/FFTBP_analysis.csv};
    \addplot[FFTBP_flooding,only marks, error bars/.cd, y dir=both, y explicit, error bar style={line width=1pt}, error mark options={rotate=90,FFTBP_flooding,mark size=1.5pt, line width=1pt}] table[x={log2N}, y={mean_iters_flooding}, y error plus={q75_iters_flooding}, y error minus={q25_iters_flooding}, col sep=comma] {numerical_results/FFTBP_analysis.csv};
    \legend{layered \\ flooding \\}
    \addplot[domain=2:13, samples=100, thick, color=gray, dotted ] {2.427 + 1.770 * ln(x)};
    \addplot[domain=2:13, samples=100, thick, color=gray, dotted ] {2.382 + 2.764 * ln(x)};
    \coordinate (legendpos) at (rel axis cs: 0.44,0.025); 
    \node[draw, fill=white, anchor=south west, font=\footnotesize, minimum width = 0em, inner sep=2pt] at (legendpos) {
        \begin{tikzpicture}
            \draw[dotted, thick, gray] (1,0) -- (4,0); %
            \node[right] at (4.1,0) {\textcolor{black}{log-regression}}; %
        \end{tikzpicture}
    };
    \nextgroupplot[ %
    width=\linewidth, %
    height=0.55\linewidth,
    align = left,
    grid=major, %
    grid style={gray!30}, %
    xlabel= ${\log_2 N}$,
    xlabel style={yshift=2pt, xshift=0pt},
    ylabel= $\log_{10} \mleft( \textrm{mean error} \mright)$,
    ymax = -0.5,
    ymin = -12.5,
    xmin = 2.8,
    xmax = 12.2,
    enlarge x limits=false,
    enlarge y limits=false,
    ytick = {-2,-5,-8,-11},
    line width=1pt,
	  legend style={font=\footnotesize, cells={align=left}, anchor=south east, at={(0.99,0.15)}},
    legend cell align={left},
    ]
    \addplot[black,only marks, mark=o, error bars/.cd, y dir=both, y explicit, error bar style={line width=1pt}, error mark options={rotate=90,black,mark size=1pt, line width=1pt}] table[x={log2N}, y={mean_var_rel_err_flooding}, y error plus={q75_var_rel_err_flooding}, y error minus={q25_var_rel_err_flooding}, col sep=comma] {numerical_results/FFTBP_analysis.csv};
    \addplot[black,only marks, mark=-, mark options={line width=1.5pt}, error bars/.cd, y dir=both, y explicit, error bar style={line width=1pt}, error mark options={rotate=90,black,mark size=1pt, line width=1pt}] table[x={log2N}, y={mean_mu_abs_err_flooding}, y error plus={q75_mu_abs_err_flooding}, y error minus={q25_mu_abs_err_flooding}, col sep=comma] {numerical_results/FFTBP_analysis.csv};
    \legend{$\textrm{tr}\mleft( \big| \bm{\Sigma}_n - \hat{\bm{\Sigma}}_{n} \big| ./ \bm{\Sigma}_n \mright)$ \\ $\left\| \bm{\mu}_n - \hat{\bm{\mu}}_{n} \right\|_1$ \\} 
    \addplot[FFTBP_flooding,only marks, mark=o, error bars/.cd, y dir=both, y explicit, error bar style={line width=1pt}, error mark options={rotate=90,FFTBP_flooding,mark size=1.5pt, line width=1pt}] table[x={log2N}, y={mean_var_rel_err_flooding}, y error plus={q75_var_rel_err_flooding}, y error minus={q25_var_rel_err_flooding}, col sep=comma] {numerical_results/FFTBP_analysis.csv};
    \addplot[FFTBP_layered,only marks, mark=o, error bars/.cd, y dir=both, y explicit, error bar style={line width=1pt}, error mark options={rotate=90,FFTBP_layered,mark size=1.5pt, line width=1pt}] table[x={log2N}, y={mean_var_rel_err_layered}, y error plus={q75_var_rel_err_layered}, y error minus={q25_var_rel_err_layered}, col sep=comma] {numerical_results/FFTBP_analysis.csv};
    \addplot[FFTBP_flooding,only marks, mark=-, mark options={line width=2pt}, error bars/.cd, y dir=both, y explicit, error bar style={line width=1pt}, error mark options={rotate=90,FFTBP_flooding,mark size=1.5pt, line width=1pt}] table[x={log2N}, y={mean_mu_abs_err_flooding}, y error plus={q75_mu_abs_err_flooding}, y error minus={q25_mu_abs_err_flooding}, col sep=comma] {numerical_results/FFTBP_analysis.csv};
    \addplot[FFTBP_layered,only marks, mark=-, mark options={line width=2pt}, error bars/.cd, y dir=both, y explicit, error bar style={line width=1pt}, error mark options={rotate=90,FFTBP_layered,mark size=1.5pt, line width=1pt}] table[x={log2N}, y={mean_mu_abs_err_layered}, y error plus={q75_mu_abs_err_layered}, y error minus={q25_mu_abs_err_layered}, col sep=comma] {numerical_results/FFTBP_analysis.csv};
  \end{groupplot}
\end{tikzpicture}
    \caption{Analysis of the \ac{GaBP} algorithm (layered and flooding schedule) on the FFT factor graph. Top: number of required \ac{BP} iterations for convergence. Bottom: logarithmic mean errors of the absolute means and the relative variances between the \ac{GaBP} beliefs and the ground truth. The plots show the mean and 25/75 percentiles over 100 randomly generated trial distributions.} 
  \label{fig:FFTBP_analysis}
\end{figure}

Fig.~\ref{fig:FFTBP_analysis} compares the layered and flooding \ac{GaBP} schedules for different \ac{FFT} sizes~$N$ regarding latency and accuracy. The top plot shows the number of layered iterations required for convergence,
revealing that the flooding schedule consistently converges faster. 
Note that the flooding schedule involves more message updates per iteration, i.e., the computational complexity is higher. %
For both schedules, the number of layered iterations scales logarithmically with $\log_2 N$, which addresses an open question in~\cite{storkey_generalised_nodate}. 

The bottom plot in Fig.~\ref{fig:FFTBP_analysis} compares \ac{GaBP} against the exact \ac{DFT} baseline of Sec.\ref{subsec:Gaussian_DFT_Uncertainty_propagation}. The flooding schedule achieves machine-precision accuracy for the means, while the layered schedule exhibits larger errors. Since the guaranteed exactness of the mean for converged \ac{GaBP} is invariant of the underlying schedule~\cite{li_fixed_2019}, we conjecture that numerical reasons cause the degraded precision of \ac{GaBP} with layered scheduling. Surprisingly, \ac{GaBP} also provides highly accurate variance estimates that improve with increasing~$N$. In addition, we found the variance errors to be unbiased for the particular graphs considered in this work, disproving the common folklore~\cite{storkey_generalised_nodate, loeliger_introduction_2004} that \ac{GaBP} systematically underestimates the variances.\footnote{To the authors knowledge, this is only proven for special cases in~\cite{weiss_correctness_2001}.}

\subsection{Symbol Detection for Noisy ISI Channels}
We consider the transmission of a discrete-time sequence~$\bm{\xt}$ with independently and uniformly sampled \ac{BPSK} symbols~${\xt_n \in \{\pm 1 \}}$, impaired by linear \ac{ISI} and additive white Gaussian noise~\cite{proakis_digital_2007}. The received signal
\begin{equation*}
    \bm{y}^\textrm{t} = \bm{\xt} * \bm{h}^\textrm{t} + \bm{n}^\textrm{t}, \quad \bm{n}^\textrm{t} \sim \mathcal{N}\mleft( \underline{\bm{n}}^\textrm{t}: \bm{0}_{2N}, \sigma^2 \bm{I}_{2N} \mright)\label{eq:ISIchannel_time}
\end{equation*}
is modeled as a linear convolution of the information sequence~$\bm{\xt}$ with the impulse response $\bm{h}^\textrm{t}=(0.04, -0.05,\allowbreak 0.07,\allowbreak -0.21,\allowbreak -0.5,\allowbreak 0.72,\allowbreak 0.36,\allowbreak 0.0,\allowbreak 0.21,\allowbreak 0.03, 0.07)^\text{T}$ of the \ac{ISI} channel.
While the discrete prior~${p(\bm{\xt})}$ is conveniently expressed in the time domain,
the likelihood factorizes more efficiently in the frequency domain:
\begin{align*}
     p(\bm{y}^\textrm{f} | \bm{\xf}) &\phantom{:}= \prod\limits_{n=1}^N \mathcal{N}\mleft(\underline{u}^\textrm{f}_n : \dunderline{\left(\frac{y^\textrm{f}_n}{h^\textrm{f}_n} \right)}, N \frac{\sigma^2}{|h^\textrm{f}_n|} \bm{I}_{2}  \mright), \; \bm{y}^\textrm{f} := \bm{\WDFT} \bm{y}^\textrm{t}, \label{eq:ISIchannel_likelihood_freq}
\end{align*}
where $\bm{\xt}$ and $\bm{h}^\textrm{t}$ are zero-padded before applying the \acp{DFT} ${\bm{\xf}=\bm{\WDFT}\bm{\xt}}$ and ${\bm{h}^\textrm{f}=\bm{\WDFT}\bm{h}^\textrm{t}}$ to ensure equivalence between linear and circular convolution.
This enables an efficient representation of~${p(\bm{\xt}|\bm{y}^\textrm{t})}$  as in~\eqref{eq:likelihood_in_freq}, enabling the EP-FFT algorithm to perform approximate \ac{MAP} detection.
In this scenario, ${p(\bm{y}^\textrm{f} | \bm{\xf})}$  is already Gaussian, and the \ac{EP} method only needs to determine the parameters~${\bm{\gamma}_n}$ and ${\bm{\lambda}_n}$ for the Gaussian approximation in the time domain.

\begin{figure}[tb]
\setlength{\abovecaptionskip}{-3pt}
\centering
  \begin{tikzpicture}
  \begin{axis}[ %
    width=0.95\linewidth, %
    height=0.66\linewidth,
    align = left,
    grid=major, %
    grid style={gray!30}, %
    xlabel= SNR (dB),
    xlabel style={yshift=3pt, xshift=0pt},
    ylabel= SER,
    ymode=log,
    ymax = 0.2,
    ymin = 0.0001,
    xmin = 0.0,
    enlarge x limits=false,
    enlarge y limits=false,
    line width=1pt,
	  legend style={font=\footnotesize, cells={align=left}, anchor=south west, at={(0.04,0.04)}},
    legend cell align={left},
    ]
    \addplot[ZF, dashed, smooth] table[x={SNR (dB)}, y={ZF}, col sep=comma] {numerical_results/symbol_detection.csv};
    \addplot[MMSE_GaussPrior, smooth] table[x={SNR (dB)}, y={LMMSE FIR}, col sep=comma] {numerical_results/symbol_detection.csv};
    \addplot[FFTBP_flooding, smooth] table[x={SNR (dB)}, y={FFTEP}, col sep=comma] {numerical_results/symbol_detection.csv};
    \addplot[EPDFT, only marks, mark=o] table[x={SNR (dB)}, y={DFTEP}, col sep=comma] {numerical_results/symbol_detection.csv};
    \addplot[MAP, dotted, smooth] table[x={SNR (dB)}, y={MAP}, col sep=comma] {numerical_results/symbol_detection.csv};
    \legend{ZF \\ LMMSE \\ EP-FFT \\ EP-DFT \\ MAP \\}
    \end{axis}
\end{tikzpicture}
    \caption{\ac{SER} over the \ac{SNR} of different detection algorithms for \ac{BPSK} transmission over an \ac{ISI} channel with block length ${K=1000}$.} 
  \label{fig:SERvsSNR_symbol_detection}
\end{figure}
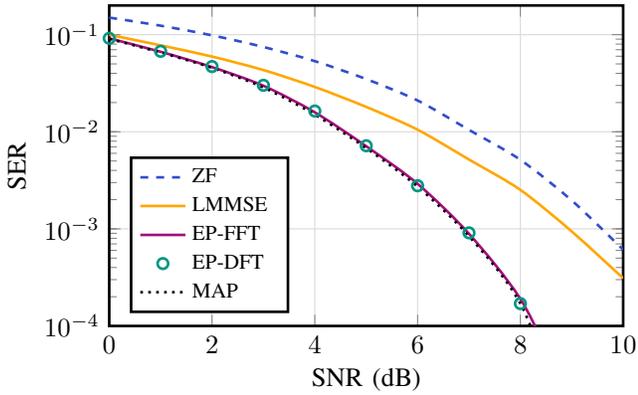

Fig.~\ref{fig:SERvsSNR_symbol_detection} shows the \ac{SER} over the \ac{SNR}. For each \ac{SNR}, we transmit $100$ randomly sampled information sequences~$\bm{\xt}$ of length ${K=1000}$ that are zero-padded to length~${N=1024}$.
With ${L=4}$ \ac{EP} iterations, both the EP-DFT and EP-FFT algorithms perform close to exact \ac{MAP} detection.
The \ac{GaBP} in the EP-FFT algorithm always converges, providing the exact Gaussian means, but with significantly reduced complexity compared to the EP-DFT baseline. 
Compared to the classical \ac{ZF} and \ac{LMMSE} equalizers\footnote{From \ac{EP} perspective, the \ac{ZF} detector ignores the discrete priors, whereas the \ac{LMMSE} detector is equivalent to $1$ iteration of the EP-DFT algorithm.}~\cite{proakis_digital_2007}, the EP-FFT algorithm achieves a $2\,\textrm{dB}$ \ac{SNR} gain at an \mbox{\ac{SER}$\;=10^{-3}$}.

\subsection{Estimation of a Multi-target Radar Channel} 
We consider an \ac{OFDM} system with ${N=1024}$ \ac{BPSK}-modulated subcarriers in a channel with multiple dominant reflectors leading to a sparse power-delay profile, e.g., in the context of a multistatic joint communication and sensing scenario.
Each component of the impulse response~${\bm{h}^\textrm{t} \in \mathbb{C}^N}$ follows a \ac{GM}
\begin{equation}
    p(h_n^\textrm{t}) = s \mathcal{N}\big( \underline{h}^\textrm{t}_n: \bm{0}_2, \bm{I}_2 \big) + (1-s) \mathcal{N}\big( \underline{h}^\textrm{t}_n: \bm{0}_2, 10^{-2}\bm{I}_2 \big), \label{eq:radar_channel}
\end{equation}
with sparsity ${s=0.01}$.
In the frequency domain, we assume uncertainty in the transmitted \ac{BPSK} symbols~$\bm{\xf}$, expressed via \acp{LLR} ${L_n} {:= \log p(\xf_n = +1)} - {\log p(\xf_n = -1) }$.
Given noisy observations ${y_n^\textrm{f} = \xf_n h^\textrm{f}_n + n^\textrm{f}}$ with ${\bm{h}^\textrm{f} := \WDFT \bm{h}^\textrm{t}}$ and ${n^\textrm{f} \sim \mathcal{N}\mleft( \underline{n}^\textrm{f}: \bm{0}_2, \sigma^2 \bm{I}_2 \mright)}$, our goal is to estimate~$\bm{h}^\textrm{t}$ in the time domain.
Since the prior~$p(\bm{h}^\textrm{t})$ and the likelihood~$p(\bm{y}^\textrm{f}|\bm{h}^\textrm{f})$ are non-Gaussian \acp{GM}, they are both approximated using the \ac{EP} method.

\begin{figure}[tb]
\setlength{\abovecaptionskip}{-3pt}
\centering
  \begin{tikzpicture}
\pgfplotsset{
compat=1.11,
legend image code/.code={
\draw[mark repeat=2,mark phase=2]
plot coordinates {
(0cm,0cm)
(0.14cm,0cm)        %
(0.28cm,0cm)         %
};%
}
} 
  \begin{axis}[ %
    width=0.95\linewidth, %
    height=0.66\linewidth,
    align = left,
    grid=major, %
    grid style={gray!30}, %
    xlabel= SNR (dB),
    xlabel style={yshift=3pt, xshift=0pt},
    ylabel= MSE,
    ymode=log,
    ymax = 0.14,
    ymin = 0.001,
    xmin = -15,
    xmax = 15,
    enlarge x limits=false,
    enlarge y limits=false,
    line width=1pt,
	  legend style={font=\scriptsize, cells={align=left}, anchor=south west, at={(-0.0024,-0.0038)}}, %
    legend cell align={left},
    ]
    \addplot[ZF, dashed, smooth] table[x={Sensing SNR (dB)}, y={c0 ZF}, col sep=comma, smooth] {numerical_results/channel_estimation.csv};
    \addplot[MMSE_noPrior, smooth] table[x={Sensing SNR (dB)}, y={c0 MMSE noAssumptionHf}, col sep=comma] {numerical_results/channel_estimation.csv};
    \addplot[MMSE_GaussPrior, smooth] table[x={Sensing SNR (dB)}, y={c0 MMSE GaussAssumptionHf}, col sep=comma] {numerical_results/channel_estimation.csv};
    \addplot[EPDFT, only marks, mark=o] table[x={Sensing SNR (dB)}, y={c0 DFTEP 4}, col sep=comma] {numerical_results/channel_estimation.csv};
    \addplot[FFTBP_flooding, smooth] table[x={Sensing SNR (dB)}, y={c0 FFTEP 4 BPflooding ctol 0.1}, col sep=comma] {numerical_results/channel_estimation.csv};
    \legend{ZF, hard decision \\ LMMSE {\scriptsize (no prior of $\bm{h}$)} \hspace{-0.9em} \\ LMMSE {\scriptsize($\mathcal{N}$-prior of $\bm{h}$)} \hspace{-0.9em} \\ EP-DFT \\ EP-FFT \\}
    \end{axis}
\end{tikzpicture}
    \caption{MSE versus SNR of the channel estimation in the time domain.} 
  \label{fig:MSEvsSNR_channel_estimation}
\end{figure}
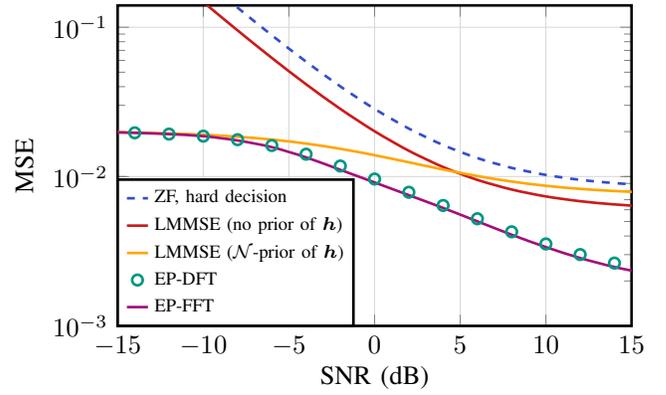

Fig.~\ref{fig:MSEvsSNR_channel_estimation} shows the estimation results for the transmission of $100$ \ac{OFDM} symbols, each with ${N=1024}$ uniformly sampled \ac{BPSK} symbols, across different \acp{SNR}. 
For each transmission, we sample a new~$\bm{h}^\textrm{t}$ according to~\eqref{eq:radar_channel} and generate \acp{LLR} from ${\mathcal{N}\mleft(L_n: \textrm{sign}(u^\textrm{f}_n)c, 2c \mright)}$.
A classical \ac{ZF} estimator performs hard symbol decisions before estimating the channel ${\hat{h}^\textrm{f}_{\textrm{ZF},n} = y_n^\textrm{f} / \textrm{sign}(L_n)}$ and transforming the result to the time domain.
We set ${c=3.25}$ such that the hard decisions yield an \ac{SER}${\; \approx 0.1}$.
Fig.~\ref{fig:MSEvsSNR_channel_estimation} also compares two \ac{LMMSE} estimators; one ignoring the prior knowledge~\eqref{eq:radar_channel}, the other approximating it as a Gaussian, similar to $1$~iteration of the EP-DFT scheme.
With ${L=4}$, the EP-DFT and EP-FFT algorithms achieve a ${5\;}$dB gain at a \ac{MSE} of ${10^{-2}}$.

\section{Conclusion}
We propose the EP-FFT algorithm for efficient approximate inference in probabilistic systems incorporating both time and frequency domain representations. 
Our numerical analysis of the underlying \ac{GaBP} algorithm on the FFT factor graph shows promising results: it consistently converged, with highly accurate estimates in mean and variance.
Our findings do not contradict those of~\cite{storkey_generalised_nodate}, where \ac{GaBP} was analyzed for missing data recovery with extreme variances (either $0$ or $\infty$). Instead, our results highlight its potential in noisy scenarios with finite variances, as typically encountered in communications and signal processing applications,
and we are optimistic about its impact in real-world applications.

\end{document}